\documentclass[conference]{IEEEtran}
\IEEEoverridecommandlockouts
\usepackage{cite}
\usepackage{amsmath,amssymb,amsfonts}
\usepackage{graphicx}
\usepackage{tabularx}
\usepackage{textcomp}
\usepackage{listings}
\usepackage{xcolor}
\definecolor{light-gray}{gray}{0.95}
\usepackage{algorithm}
\usepackage{algpseudocode}
\def\BibTeX{{\rm B\kern-.05em{\sc i\kern-.025em b}\kern-.08em
    T\kern-.1667em\lower.7ex\hbox{E}\kern-.125emX}}
\usepackage{enumitem}
\newlist{requirements}{enumerate}{1}
\setlist[requirements,1]{wide=0pt, label=R\arabic*, widest=\textbf{R9}, font=\bfseries, leftmargin=*, ref=\textbf{R\arabic*}}
\usepackage{cleveref}
\crefname{requirementsi}{Requirement}{requirements}
\Crefname{requirementsi}{Requirement}{Requirements}

\begin{document}

\title{Representing Timed Automata and Timing Anomalies of Cyber-Physical Production Systems in Knowledge Graphs\\
}

\author{
\IEEEauthorblockN{Tom Westermann}
\IEEEauthorblockA{\textit{Institute of
Automation Technology},\\
\textit{Helmut-Schmidt-University}\\
Hamburg, Germany \\
tom.westermann@hsu-hh.de}
\and
\IEEEauthorblockN{Milapji Singh Gill}
\IEEEauthorblockA{\textit{Institute of
Automation Technology},\\
\textit{Helmut-Schmidt-University}\\
Hamburg, Germany \\
milapji.gill@hsu-hh.de}
\and
\IEEEauthorblockN{Alexander Fay}
\IEEEauthorblockA{\textit{Institute of
Automation Technology},\\
\textit{Helmut-Schmidt-University}\\
Hamburg, Germany \\
alexander.fay@hsu-hh.de}
}

\maketitle

\begin{abstract}
Model-Based Anomaly Detection has been a successful approach to identify deviations from the expected behavior of Cyber-Physical Production Systems. 
Since manual creation of these models is a time-consuming process, it is advantageous to learn them from data and represent them in a generic formalism like timed automata.
However, these models - and by extension, the detected anomalies - can be challenging to interpret due to a lack of additional information about the system. 
This paper aims to improve model-based anomaly detection in CPPS by combining the learned timed automaton with a formal knowledge graph about the system. 
Both the model and the detected anomalies are described in the knowledge graph in order to allow operators an easier interpretation of the model and the detected anomalies.
The authors additionally propose an ontology of the necessary concepts. 
The approach was validated on a five-tank mixing CPPS and was able to formally define both automata model as well as timing anomalies in automata execution.
\end{abstract}


\section{Introduction} \label{introduction}
In the field of production systems, research agendas like CPS \cite{Lee.2008} promote the investigation into Cyber-Physical Production Systems (CPPS), with the aim of increasing flexibility and enabling self-surveillance and self-diagnosis capabilities \cite{O.Niggemann.2015}. 
To achieve these goals, the use of timing information -- i.e. the time span that passes between events in a production sequence -- is a promising but often overlooked source of information \cite{O.Niggemann.2015}. 
In order to detect anomalies in such systems - usually deviations from the expected system behavior - two main approaches exist: 
On one hand, there are phenomenological approaches, in which observations are directly classified as correct or anomalous.
On the other hand, there are model-based approaches, in which a model is used to simulate the normal behavior of a system and to compare this to the observed behavior. For complex distributed systems, the latter approach is considered to be preferable \cite{O.Niggemann.2015}.

However, due to the complex and ever-changing nature of CPPS, manually creating these models has proven to be ineffective \cite{O.Niggemann.2015}. 
Instead, multiple approaches exist to learn models of timing behavior from the systems' observed data sequences \cite{A.Maier.2011, Hranisavljevic.2020, Verwer.2010}. These models, often in the form of timed automata, can then be used in different applications, including anomaly detection \cite{A.Maier.2015} and subsequent diagnosis \cite{Bunte.2019}.
While this kind of online data generated during operation describes the real-time status of the system, for CPPS additional sources of information exist, such as engineering knowledge and configuration data. 
These are generated during the engineering of the system and describe the physical part of a CPS \cite{Monostori.2014}.
Without this engineering knowledge, the interpretation of the behavior models is considerably more difficult as they can be hard to relate to the system's actual behavior. 
Especially for the diagnosis of faults that caused the observed anomalies, this information is essential \cite{O.Niggemann.2015}.

While this information is available in engineering documents, users of data-driven approaches often suffer from unavailable meta data \cite{VogelHeuser.2021}, since it is heterogeneous and stored in (possibly inaccessible) data silos \cite{Jirkovsky.2017}.
In the field of CPPS, knowledge graphs and ontologies have been successful in integrating heterogeneous engineering documents into a well-defined shared conceptualization\cite{Jirkovsky.2017}.
In these applications, the knowledge graph contains facts about the relationships between entities, while the ontology provides the data schema of allowed entity classes, relationship types and their formal semantics.

Therefore, the approach presented in this work aims to improve the interpretability of model-based anomaly detection in CPPS by using prior engineering information. 
It combines learned timed automata with a formal knowledge base that describes the physical structure as well as engineering information  of the production system. 
Both the learned automaton as well as any detected timing anomalies are described in relation to the knowledge graph. 

The knowledge graph used to connect prior knowledge to the learned model must therefore fulfill the following requirements:
\begin{requirements}
    \item Formally describe both the physical as well as the behavioral part of a CPPS in a knowledge graph \label{rq:cyberPhysical}
    \item Formally describe all concepts needed to define learned timed automata (i.e. states $S$, events $\Sigma$, transitions $T$, and timing constraints $\Delta$ \cite{Alur.1994}) with regard to the knowledge graph from \cref{rq:cyberPhysical} \label{rq:defineStates}
    \item Formally describe detected anomalies with regard to the knowledge graph from \cref{rq:cyberPhysical} and the timed automaton from \cref{rq:defineStates} \label{rq:defineAnomalies}
\end{requirements}

An approach that fulfills these requirements is described in Section \ref{Approach} and validated on an exemplary use case in Sections \ref{useCase} and \ref{validation}. 
It produces an enriched model that allows the operator an easier understanding of both the automaton as well as the detected anomalies. 
Additionally, linking up the detected anomalies to engineering knowledge about the system offers possibilities to extend this approach towards diagnosis algorithms from the field of informed machine learning \cite{Rueden.29.03.2019}.

\section{Related Work}\label{relatedWorks}
As the presented approach combines a knowledge graph of a system with learned timed automata, this section will provide a short overview of relevant works from both areas.

In the field of knowledge representation, Hildebrandt et al. present a domain-expert-centric approach to ontology design of CPPS, focusing on 
industry standards as ontology design patterns \cite{Hildebrandt.2020}. The approach is validated on an industrial use case encompassing system structure, processes, states, and data elements and has a strong focus on ontology reuse. 

Schneider et al. \cite{Schneider.2019} follow a similarly modular approach to ontology design, which they use to formally define a modular domain ontology that describes the cyber and physical part of an automation system. The knowledge base was used to verify different control logic types in cyber- and physical parts of the system in an incremental manner.
Of particular interest to this work is the description of the systems' behavior 
using UML state machines.  
The authors of this work used a similar approach to ontology building to interpret heterogeneous observed data from an OPC UA server through the lens of a semantically enriched process description\cite{T.Westermann.2022}. In a separate work, they explored the use of knowledge graphs to reduce the manual effort in data mining approaches\cite{Gill.21.07.2023}. 

In the field of algorithmic learning of a system’s behavior model, multiple approaches exist. 
Verwer et al \cite{Verwer.2010} describe an algorithm called real-time identification from positive data (called RTI+ in short), which extracts a timed automaton from a system's observed events by creating a Prefix-Tree-Acceptor (PTA) and subsequently merging its leafs based on a compatibility criterion.

Maier et al. \cite{A.Maier.2011} introduce the Bottom-Up Timed Automaton Learning Algorithm (BUTLA), which also employs PTAs, but uses different strategies to learn timing behavior and perform state merging. The authors further present an anomaly detection algorithm (ANODA) that compares previously unseen event sequences to a timed automaton that was learned from a CPPS. 
By providing the timed automaton with a set of functions that calculate signal value changes within states, Vodenčarević. et. al \cite{Vodenčarević.2011} extend BUTLA -- which was designed with discrete systems in mind -- towards hybrid timed automata.

Maier et al. \cite{A.Maier.2015} further describe the Online Timed Automaton Learning Algorithm (OTALA), which uses state vectors to encode states of production systems. Unlike RTI+ and BUTLA, it does not employ state merging and can be applied in an online manner. 
Bunte et. al \cite{A.Bunte.2019} also tackled the problem of difficulty in interpreting the resulting models by presenting both an online and offline algorithm to detect production cycles in automata that were learned using OTALA. 
Hranisavljevic et. al \cite{Hranisavljevic.2020} present an approach that combines a state vector formed from a system's discrete variables with a deep learning-based discretization of the system's continuous variables. 
It is therefore an extension of algorithms like OTALA, that only take discrete variables into account. 

While algorithms like RTI+ and BUTLA use pre- and postfixes of the event trees and therefore consider states to have memory, OTALA and DENTA consider states to be memory-less. Different event sequences can therefore lead to the same state. Due to their reliance on state vectors, they instead offer a simpler interpretation of the learned behavior. 

Even though approaches exist that define a preexisting behavior model of a CPPS \cite{Schneider.2019} in a knowledge graph, presently no approaches exist that define a behavior model which was learned by means of a learning algorithm.

\section{Approach}\label{Approach}
\begin{figure}[ht]
\includegraphics[width=0.48\textwidth]{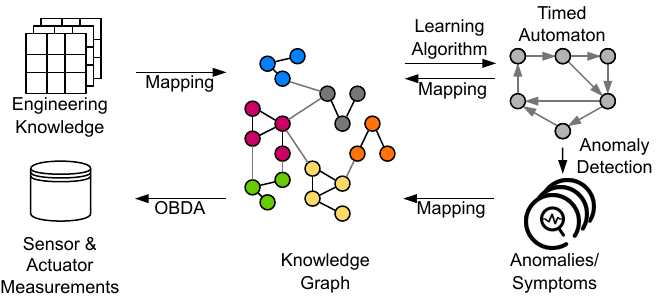}
\caption{Proposed Approach: Learning a timed automaton of a CPPS from data in a knowledge graph, using it for anomaly detection and then defining the automaton and anomalies in the knowledge graph.}
\label{fig:Approach}
\end{figure}

The following section describes an approach for the combination of a learned behavior model with prior knowledge that was generated during the engineering phase of a CPPS. 
In order to achieve this goal the approach relies on four interacting artifacts (s. Figure \ref{fig:Approach}): 
\begin{itemize}
    \item a formally defined knowledge base in the form of a graph and an associated ontology
    \item a learning algorithm that can create a timed automaton from recorded runtime data
    \item an anomaly detection algorithm that classifies unseen sequences of events
    \item mappings which define the learned timed automaton and the detected anomalies with regard to the knowledge base.
\end{itemize}

\subsection{Knowledge Graph} \label{sec:KnowledgeBase}
\begin{figure*}[t]
\includegraphics[width=1\textwidth]{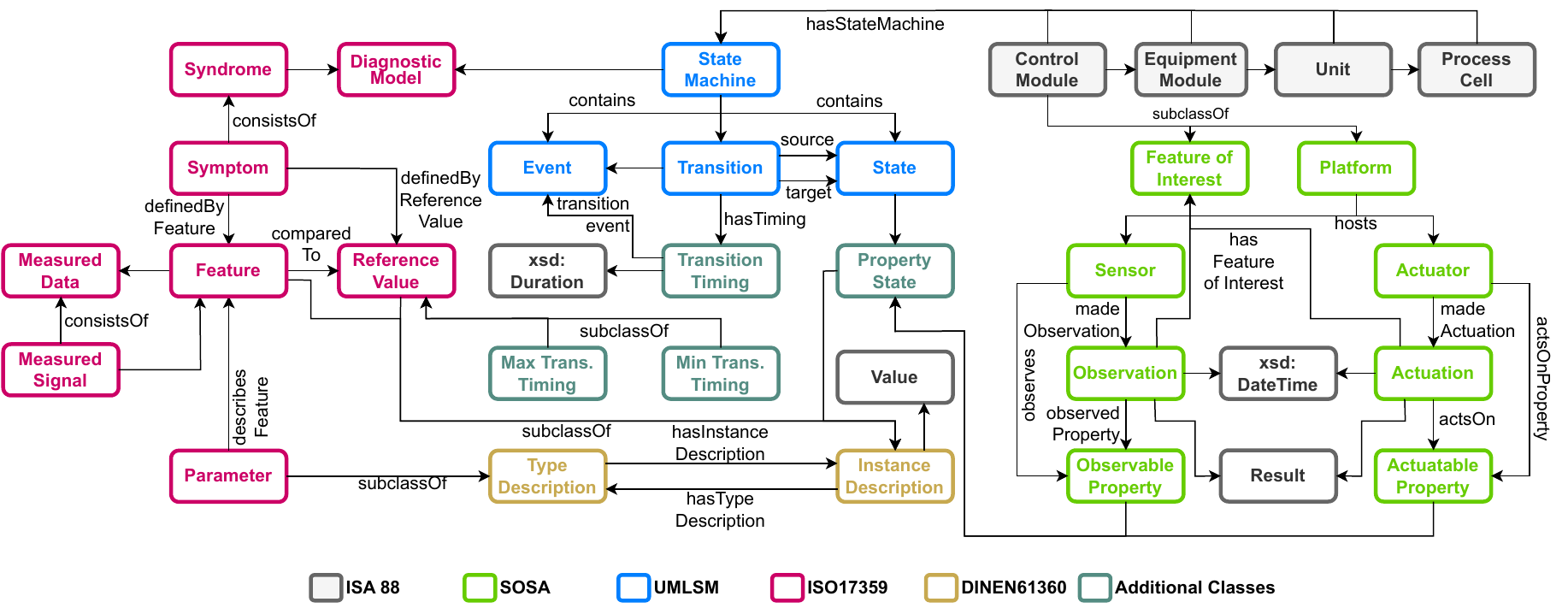}
\caption{Excerpt from the proposed Lightweight Alignment Ontology.}
\label{fig:alignmentOntology}
\end{figure*}
As described in Section \ref{introduction}, a complete knowledge graph consists of a collection of facts about entity relationships as well as an ontology that defines which classes of entities and relationships can exist. 

Therefore, in the first step,  an ontology that defines all the necessary classes and relationships to meet the requirements outlined in section \ref{introduction} is designed.

The ontology creation follows the approach outlined by Hildebrandt et al. \cite{C.Hildebrandt.2018}, which proposes using Ontology Design Patterns (ODP) based on industrial standards to model the necessary terms of the domain. In order to identify the necessary concepts of the ontology, Competency Questions (CQ) are created, that the ontology is supposed to answer. 

In this approach, the following ODPs were selected:
For \cref{rq:cyberPhysical} information about the physical structure of the CPPS is needed, including a semantically unambiguous description of the system's sensors and actuators.
The Semantic Sensor Network Ontology (SSN/SOSA) \cite{Haller.2018} is a widely applicable ontology to define concepts regarding sensors, actuators, observations and samplings. The SSN/SOSA-Ontology does not define properties that sensors can measure and instead leaves this to domain-specific ontologies. In the context of CPPS, this can be achieved by combining it with the design pattern based on the DINEN61360. If used alongside the ECLASS-vocabulary, this design pattern can be used to semantically define which kind of property a sensor observes.  

The concepts from the physical model defined in the ISA88 are widely used in the batch industry, and can be used to describe a system's hierarchy. 
However, other design patterns contain a system hierarchy as well (e.g. the CAEX ontology), which would make them equally suitable. 
Using the system hierarchy it is possible to define where a sensor or actuator is mounted. All of these design patterns can be reused from previous projects and together define the physical aspects of the CPPS according to \cref{rq:cyberPhysical}.

In order to describe the CPPS' behavior in the form of finite state automata, the UML state machine design pattern from Schneider et. al \cite{Schneider.2019} can be used. 
Since finite state automata are a tuple $(S, \Sigma, T)$, the ODP offers a possibility to define states $S$, events $\Sigma$ and transitions $T$. 
By extending it with transition timing concepts to define the timing constraints $\delta$, timed automata as per \cref{rq:defineStates} can be described as well. 
In order to describe timing anomalies according to \cref{rq:defineAnomalies}, a new ODP based on the ISO17359 was designed. It defines a wide selection of concepts regarding condition monitoring and diagnostics of machines. Of particular interest to this approach is the possibility to define symptoms, which are deviations of some measurement, compared to a reference value. These concepts are used to describe timing anomalies, in which a transition occurred, whose transition timing is outside the timing constraints $\delta$.


From these ODPs, a larger alignment ontology was created, that combines all classes from the respective ODPs into a shared conceptualization, which looks as follows: 

Control Modules from the ISA88-terminology can be considered platforms and features of interest for sensors/actuators from the SOSA ontology. Any physical model entity from the ISA88 can have an assigned state machine. The states inside these state machines can have property states, which are defined by information regarding the properties from the SOSA ontology. 
The resulting alignment ontology can be seen in Figure \ref{fig:alignmentOntology}. An instance-level modeling of these concepts can be found in Section \ref{useCase}.

All design patterns are publicly available on github\footnote{ISA88, ISO17359, DINEN61360: https://github.com/hsu-aut/}\footnote{UML State Machines: https://w3id.org/ibp/StateMachineOntology/}.

\subsection{Timed Automata Learning Algorithm} \label{sec:automataLearning}
As shown in Section \ref{relatedWorks}, there are a multitude of algorithms to learn timed automata from discrete events. 

According to Alur \cite{Alur.1994}, a timed automaton is a tuple $A = (S, S_{0}, \Sigma, T, \Delta, c)$, where:

\begin{itemize}
    \item $S$ is a finite set of states, $S_{0}\in S$ are a set of initial states,
    \item $\Sigma$ is the alphabet comprising all the relevant events $e$, 
    \item $T \subseteq S \times \Sigma \times S$ defines the set of transitions. For a transition $( s, a, s', \delta )$, the states $s,s' \in S$ are the source and destination states, $a \in \Sigma$ is the trigger event and $\delta \in \Delta$ defines the timing constraints,
    \item A set of transition timing constraints $\Delta$ with $\delta: T \rightarrow I$, $\delta \in \Delta$, where $I$ is a set of time intervals, $\delta$ defines the minimum and maximum allowed timing for each transition,
    \item A single clock $c$ is used to record the time evolution. At each transition, the clock is reset, which allows the modeling of relative time steps.
\end{itemize}

While some timed automata learning algorithms like BUTLA \cite{A.Maier.2011} or RTI+ \cite{Verwer.2010} use state merging, others (e.g. DENTA\cite{Hranisavljevic.2020} and OTALA \cite{A.Maier.2015}) represent states in the form of state vectors. Here, each state $s\in S$ has an associated signal vector $u = (io_1, io_2,..., io_n)$, which is defined over discrete variables used for training \cite{A.Maier.2015}. For CPPS, these are usually the active/inactive IO values of the system. 

Therefore, timed automata with states represented by state vectors implicitly contain the values of the discrete variables used for training. 
Due to this, they offer a solid basis for a formal description.
For the presented approach, OTALA \cite{A.Maier.2015} was chosen. 
Its objective is to determine a state-based timed automaton for the behavioral description of the system based on the observed data. 
In this variation of timed automata the set of events $\Sigma$ is comprised of any changes $e$ in the state vector $u$ (e.g. an IO variable changing its value), which then trigger transitions $T$. 
 
The learning algorithm functions in an online manner, where events are ingested as they occur. Given a state $s_i$ an event $a$ is observed, that results in a transition $e$ to state $s_j$. If $s_j$ is not part of the set of states $S$, it is added. If $s_j$ is part of $S$ already, its upper and lower timing constraints $\delta$ are updated. This process is repeated until the number of states converges. For a detailed description please refer to \cite{A.Maier.2015}. 



\subsection{Anomaly Detection Algorithm} \label{sec:anomalyDetection}
In order to detect anomalies in the systems' behavior, the Anomaly Detection Algorithm ANODA, proposed by Maier et al. \cite{A.Maier.2011}, is used. 
It employs the timed automaton from Section \ref{sec:automataLearning} to classify previously unseen event sequences into allowed and anomalous sequences. 

For this, it distinguishes between two types of anomalies:
\begin{itemize}
    \item \emph{Functional Errors}, where given a current state $s$ the observed event $e$ is not in the set of possible outgoing events, and
    \item \emph{Timing Errors}, where given a current state $s$ the observed event $e$ is allowed, but violates the learned timing constraints $\delta$ of the transition.
\end{itemize}
Algorithm \ref{alg:ANODA} describes this classification process in pseudocode.

\begin{algorithm}[H]
\caption{Discrete Anomaly Detection Algorithm ANODA} \label{alg:ANODA}
\begin{algorithmic}[1]
\Require Timed Automaton $A = (S, S_{0}, \Sigma, T, \Delta, c)$
\Require Observation $o=(e,t)$
\Ensure $s \in S$ at the beginning of the sequence
\If{exists $e \in T$ with $e = (s, a, s')$}
    \If{$t$ satisfies $\delta(e)$}
        \State{$s_{new}:=s'$}
    \Else{}
        \Return{Anomaly: Wrong Timing}
    \EndIf
\Else{}
    \Return{Anomaly: Unknown Event}
\EndIf{}
\State{\Return{$s_{new}$}}
\end{algorithmic}
\end{algorithm}

\subsection{Representing Model and Anomalies in Knowledge Base}
Once a timed automaton was learned and used to detect anomalies in the data, the results need to be connected to the prior knowledge about the system. 
This can be achieved by first combining the physical model described in \ref{sec:KnowledgeBase} with a formal description of the automaton from \ref{sec:automataLearning}, followed by defining the detected anomalies from \ref{sec:anomalyDetection} with regard to the automaton. 
\subsubsection*{Mapping Timed Automaton Model}
The timed automaton can be defined using the ODP of the UML-State-Machine-Ontology, as it describes states $S$, transitions $T$ and events $\Sigma$ of the system.
Out of the box, the design pattern is suitable to describe finite state machines. In order to extend it towards timed automata as per \cref{rq:defineStates}, a \emph{Transition Timing}-class was added, which describes the timing limits $\delta$ of a transition. 

To improve the interpretability of the learned automaton, a class \emph{Property State} is introduced. This is used to define the value of a property from the SOSA ontology for a certain state (e.g. that a certain valve is open in that state). 

Additionally, events are defined with regard to the actuations and observations in the SOSA ontology. They are therefore linked to the hardware that generated the event. 

\subsubsection*{Mapping Timing Anomalies}
In order to describe anomalies, the terminology from the ISO17359-ODP can be used, which defines terms and relationships needed for condition monitoring and diagnostics of machines. 
The state machine functions as a diagnostic model, that detects timing anomalies (called Symptoms in the standard). 
These timing anomalies are defined as transitions, whose transition timing is outside the condition timing constraints $\delta$, which are described as reference values. 
Through linking the anomalies to the well-defined states and events of the state machine, interpretation of the anomaly and subsequent identification of appropriate countermeasures is made easier. 

\section{Use Case: Five Tank Mixing-CPPS}\label{useCase}
\begin{figure}[t]
\includegraphics[width=0.48\textwidth]{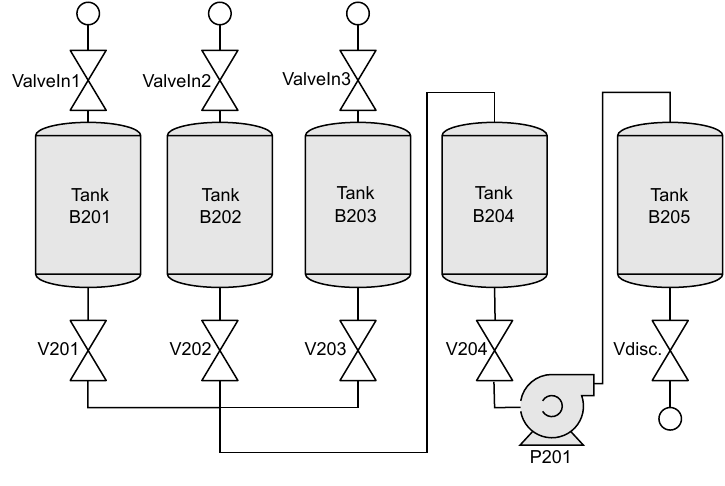}
\caption{Five-Tank Mixing Module that was considered as a use case.}
\label{fig:MixingModule}
\end{figure}

In order to validate the approach outlined in Section \ref{Approach}, the five Tank Mixing-CPPS from \cite{J.Ehrhardt.2022} was used. 
It consists of three tanks capable of storing different fluids, which can be mixed in a fourth and pumped into a fifth tank. 
During regular production, the three input tanks are sequentially filled. The fluid is then mixed inside a reservoir and pumped into a fourth tank. Finally, tank B205 is emptied and the cycle starts over. A P\&I-Diagram of the system can be seen in Figure \ref{fig:MixingModule}. 
The system contains a total of 9 actuators as well as several dozen sensors that measure properties like the filling level, fluid temperature or mass flow. Sensors and actuators record observations at a sampling rate of 100ms.

Ehrhardt et al. \cite{J.Ehrhardt.2022} present a high-fidelity simulation model of the system, in which two kinds of faults can be induced: a leakage diverts some volume flow into a separate sink, while a clogging restricts the flow through one of the pipes. 

\subsection{Knowledge Graph}
The alignment ontology described in Figure \ref{fig:alignmentOntology} was created using the selected ODPs. 
The knowledge graph was then populated with facts about the production system, which were available from engineering documents that describe the structure of the system. 
These facts contain information about the physical part of the system. This entails the systems hierarchy, as well as information about the various sensors and actuators, e.g. where those sensors are located, what kind of property they observe, as well as additional metadata. 
The facts were available in CSV- and JSON-files that were translated into rdf-triples using the mapping language \emph{RML}\cite{Dimou.2014} and the Python-based mapping interpreter \emph{SDM-RDFizer} \cite{Iglesias.2020}. 
 
To avoid inflating the knowledge graph with the high number of observations and actuations, these were stored in a separate database. At query time, triples describing the observations and actuations are created using \emph{R2RML}-mappings and the Ontology-Based Data Access-Tool \emph{Ontop} \cite{Calvanese.2016}. 

With physical information regarding sensors and actuators, observations and actuations available in the knowledge graph, a SPARQL-query can be formulated that retrieves all observations and actuations of a (sub-)system in a given time frame. 


\subsection{Learning Timed Automaton}
In the learning phase, the OTALA-Algorithm described in Section \ref{Approach} was used. 
Nine actuators - eight valves and one pump - were used to create the state vector $u = (io_1, io_2,..., io_9)$.
All actuations of five hours of undisturbed production were recorded and used to create a training set. 
Using a python implementation of the OTALA algorithm, a timed automaton was learned that represents the system's behavior. 
Due to the cycle times of the PLC code, multiple actors often change their values simultaneously. If these changes occurred in the same PLC cycle, the events were treated as one event. The transition timings were modeled using the lowest and highest observed timings as limits.
\begin{figure}[t]
\includegraphics[width=0.48\textwidth]{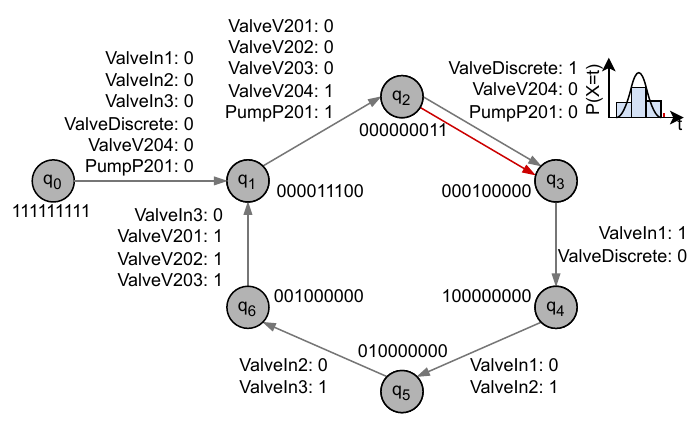}
\caption{Learned Automaton of the five tank mixing system showing state vectors, transition events and an exemplary timing distribution.}
\label{fig:LearnedTimedAutomaton}
\end{figure}

The learned automaton can be seen in Figure \ref{fig:LearnedTimedAutomaton}. It consists of one initial state and six production states that are activated in a cyclical manner. 

\subsection{Anomaly Detection}
For the anomaly detection, a test set was created where a fault was introduced in the system by clogging the pipe leading to pump P201. 
This was achieved by restricting the diameter of the pipe by a set amount, thus restricting the fluid flow through the pipe. 
In the system, this change can be detected by a slight increase in the time needed to fill Tank B204. 
The test set was used alongside the learned timed automaton of the system in the anomaly detection algorithm ANODA. 
The model is able to detect the fault as a prolonged activation time of state $q_2$. The state was active for about 127 seconds while the clogging was active compared to a maximum of 121.8 seconds in regular conditions. 

\subsection{Representing Model in Knowledge Base}
\begin{figure}[t]
\includegraphics[width=0.48\textwidth]{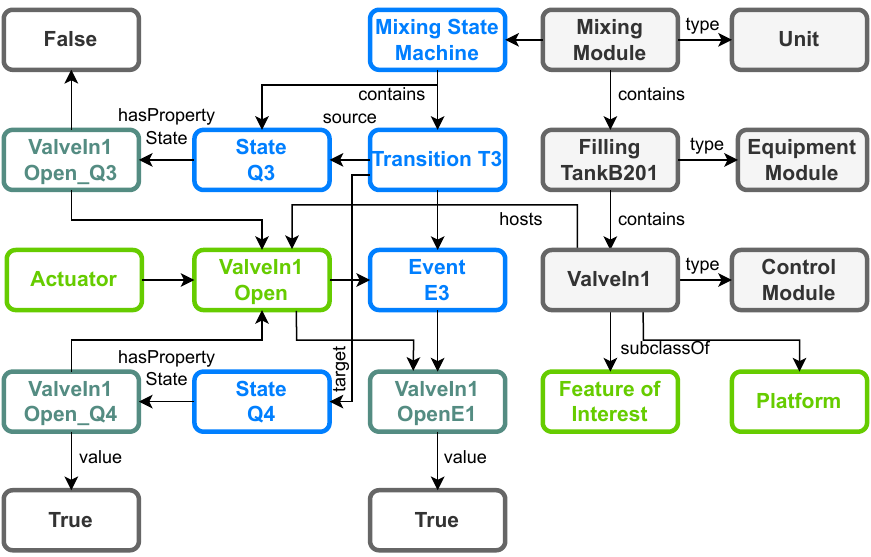}
\caption{Representation of an excerpt of the timed automaton in the knowledge graph. }
\label{fig:StateMachineInstance}
\end{figure}

\begin{figure}[b]
\includegraphics[width=0.48\textwidth]{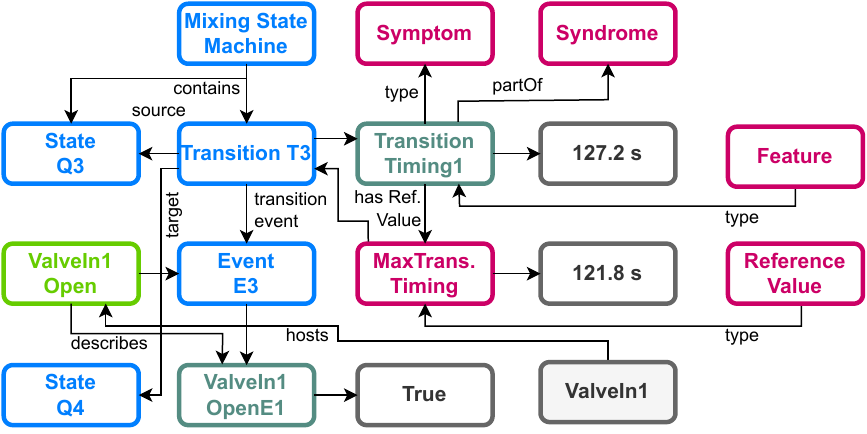}
\caption{Representation of a detected timing anomaly in the learned automaton in the knowledge graph. }
\label{fig:AnomalyInstance}
\end{figure}

\begin{table*}[t]
  \centering
  \caption{Excerpt of Competency Questions from the the Five Tank Mixing-CPPS}
  \begin{tabularx}{0.9\textwidth}{X|p{0.25\textwidth}|p{0.1\textwidth}}
    \textbf{Competency Question} & \textbf{Answer} & \textbf{Requirement} \\
    \hline
    Which sensors are part of Tank B201? & tank\_B201.level, B201\_isFull & \cref{rq:cyberPhysical}\\
    Which actuators are part of the Mixing Module? & Valves V201, V202, V203, ... & \cref{rq:cyberPhysical}\\
    Where is sensor Tank\_B201.level mounted? & Tank\_B201 & \cref{rq:cyberPhysical}\\
    What property does Tank\_B201.level measure? & Filling Level of Tank\_B201 & \cref{rq:cyberPhysical}\\
    How many states does the State Machine of the Mixing Module contain? & Filling Level of Tank\_B201 & \cref{rq:cyberPhysical}\\
    What was the state of Valve V204 in state \emph{q2}? & Open & \cref{rq:defineStates}\\
    In which state was ValveV204 open and Pump P201 turned on? & State \emph{q2} & \cref{rq:defineStates}\\
    Which events are allowed in state \emph{q3}? & ValveIn1: Close, ValveIn2: Open & \cref{rq:defineStates}\\    
    Between which two states was a timing anomaly observed? & Between States \emph{q2} \& \emph{q3} & \cref{rq:defineAnomalies}\\
    By how many seconds was the anomaly outside the max. transition time? & By 5.2 Seconds & \cref{rq:defineAnomalies}\\
    Which timing anomalies were part of a syndrome? & Trans.Timing1 and Trans.Timing2 & \cref{rq:defineAnomalies}\\     
    Which event should have occurred when the timing anomaly was observed? & ValveDiscrete: 1, \newline ValveV204: 0, \newline PumpP201: 0 & \cref{rq:defineAnomalies}\\           
    ... & ... & ... \\
  \end{tabularx}
\label{tab:CompetencyQuestions}
\end{table*}

In order to represent the learned model in the knowledge graph, the automaton was mapped to RDF-triples. 
Both the property states and the event description are connected to an Actuator and its associated value. 
In this way, the knowledge graph contains information of the current configuration of a plant while a state is active. 
State $q_2$, for example, is described by nine property states, which tell us that pump P201 is turned on, while only valve V204 is open. This should correspond to a fluid transfer from the reservoir to tank B204. 

The state is ended by an event $e_3$. It is described by three event descriptions, which contain the information that Pump P201 was turned off, Valve V204 was closed and Valve Discrete was opened. 
A small excerpt of the knowledge graph can be seen in Figure \ref{fig:StateMachineInstance}.

\subsection{Representing Anomalies in Knowledge Base}
In order to represent the detected anomalies, a similar approach was followed. 
Anomalies were described using the classes and relationships from the ISO17359 ODP. 
Figure \ref{fig:AnomalyInstance} describes an example of a symptom, where a transition timing occurred, that was larger than the learned maximum transition timing.

\section{Validation}\label{validation}
With the learned Timed Automaton, the detected timing anomalies and the physical model of the CPPS now represented in the knowledge graph, they can now be evaluated against the requirements outlined in Section \ref{introduction}.
For this, the requirements were translated into a set of competency questions, that define both necessary questions as well as expected answers. 
In order to test the knowledge graph against these questions, various SPARQL-queries were written that cover many different information retrieval operations.
Using these SPARQL-queries it is now possible for an operator to:
\begin{itemize}
    \item Search the state machine of a module
    \item Search for states that have some characteristics (e.g. certain combinations of actuator states)
    \item Access supplementary information about a detected anomaly (e.g. the timing deviation or the event that was observed) 
    \item Create a subgraph that describes an observed syndrome alongside information about the physical system. 
\end{itemize}

A more exhaustive list of competency questions and associated answers can be seen in Table \ref{tab:CompetencyQuestions}. 

In comparison to the 'plain' representation of timing anomalies, the representation in the knowledge graph is enriched with information regarding the anomalies' context. 
This context can be very valuable in enabling further understanding of the cause of the anomaly. 
In case of an anomaly, the enriched representation can not only provide the event and transition timing that was observed, but also provide a set of events and transition timings that would have been acceptable. 
Providing this data enables the operator to make a more informed assessment of the nature and severity of the anomalies and helps determine their causes.


\section{Conclusion \& Future Research}
Model-based anomaly detection and diagnosis have shown to be successful approaches towards enabling self-surveillance and self-diagnosis capabilities of CPPS. 
However, the models needed for these capabilities are difficult to create manually, which is why they are often learned instead. 
While learning timed automata of CPPS have proven successful in mitigating the high effort needed for their creation, it comes at the cost of a more difficult interpretation of its states and subsequent anomaly detection results. 
In this paper, it was shown that by describing the timed automata and timing anomalies in a knowledge graph and combining it with further engineering knowledge about the system, the interpretability of these models can be improved considerably. 
The approach was validated on a small example CPPS. Since the representation approach is independent of the system size, it is expected to generalize well to larger systems. The timed automaton learning algorithm however underlies limitations with regard to larger systems, that need to be further investigated. 

For future research, the combination of a learned timed automaton with an extensive collection of prior knowledge in an accessible form can be used to apply numerous diagnosis algorithms. 
Due to ontologies built-in support for logical rules, extending the existing knowledge base with a causality model would be a possible approach. 
Additionally, since a syndrome of multiple timing anomalies can be described in a rich subgraph, approaches from the field of graph classification and graph embedding now become possible.


\section*{Acknowledgment}
This work has been partially supported and funded by the German Federal Ministry of Education and Research (BMBF) for the project ”Time4CPS - A Software Framework for the Analysis of Timing Behaviour of Production and Logistics Processes” under the contract number 01IS20002.
This work has been partially funded by dtec.bw – Digitalization and Technology Research Center of the Bundeswehr as part of the project ProMoDi. dtec.bw is funded by the European Union – NextGenerationEU.

\bibliographystyle{IEEEtran}
\bibliography{main.bib}

\end{document}